\def\year{2020}\relax
\documentclass[letterpaper]{article} % DO NOT CHANGE THIS
\usepackage{aaai20}  % DO NOT CHANGE THIS
\usepackage{times}  % DO NOT CHANGE THIS
\usepackage{helvet} % DO NOT CHANGE THIS
\usepackage{courier}  % DO NOT CHANGE THIS
\usepackage[hyphens]{url}  % DO NOT CHANGE THIS
\usepackage{graphicx} % DO NOT CHANGE THIS
\usepackage{graphicx}  % DO NOT CHANGE THIS
\usepackage{amsthm,amssymb,amsmath,amsfonts}
\usepackage{mathtools}
\usepackage{xspace,url}
\usepackage[shortlabels]{enumitem}
\usepackage{graphicx}
\usepackage{multirow,booktabs,colortbl,tabularx}
\usepackage[style=english]{csquotes}
\usepackage{MnSymbol}
\usepackage{etoolbox}
\usepackage{subcaption}
\usepackage{todonotes}
\usepackage{tikz}
\usepackage{wasysym}
\usetikzlibrary{arrows,calc}
\usepackage{xfrac}
\usepackage{setspace}
\usepackage{pgfplots}
\newcommand{\MC}{MC\xspace}
\newcommand{\WMC}{WMC\xspace}
\newcommand{\SAT}{SAT\xspace}
\newcommand{\DNF}{DNF\xspace}
\newcommand{\CNF}{CNF\xspace}
\newcommand{\sSAT}{\#SAT\xspace}
\newcommand{\sDNF}{\#DNF\xspace}
\newcommand{\sCNF}{\#CNF\xspace}

\newcommand{\minw}{\mathsf{minW}\xspace}
\newcommand{\maxw}{\mathsf{maxW}\xspace}
\newcommand{\PTime}{\textsc{P}\xspace}

\newcommand{\NP}{\textsc{NP}\xspace}

\newcommand{\sharpP}{\textsc{\#P}\xspace}
\title{Learning to Reason: Leveraging Neural Networks \\ for Approximate DNF Counting}
\author{Ralph Abboud, {\.I}smail {\.I}lkan Ceylan, Thomas Lukasiewicz\\ 
Department of Computer Science\\ University of Oxford, UK \\
\{ralph.abboud, ismail.ceylan, thomas.lukasiewicz\}@cs.ox.ac.uk
}

\begin{document}

\maketitle
\begin{abstract}

Weighted model counting (WMC) has emerged as a prevalent approach for probabilistic inference. In its most general form, WMC is \sharpP-hard. %and, as a result, solving real-world WMC instances is intractable. 
Weighted \DNF counting (weighted \sDNF) is a special case, where approximations with probabilistic guarantees are obtained in $O(nm)$, where $n$ denotes the number of variables, and $m$ the number of clauses of the input DNF, but this is not scalable in practice. 
In this paper, we propose a \emph{neural model counting} approach for weighted \sDNF that combines approximate model counting with deep learning, and accurately approximates model counts in linear time when width is bounded. We conduct experiments to validate our method, and show that our model learns and generalizes very well to large-scale \sDNF instances.
\end{abstract}

\section{Introduction}
Propositional \emph{model counting~(\MC)}, or \emph{\sSAT}, is the task of counting the number of satisfying assignments  for a given propositional formula~\cite{Gomes09}.  \emph{Weighted model counting~(\WMC), or weighted \sSAT,} additionally incorporates a \emph{weight function} over the set of all possible assignments. Offering an elegant formalism for encoding various probabilistic inference problems, \WMC is a unifying approach for inference in a wide range of probabilistic models. In particular, \emph{probabilistic graphical models}~\cite{Koller-PGM}, \emph{probabilistic planning}~\cite{Domshlak07}, \emph{probabilistic logic programming}~\cite{ProbLog}, \emph{probabilistic databases}~\cite{Suciu-PDBs}, and \emph{probabilistic ontologies}~\cite{BCL-AAAI17} can greatly benefit from advances in \WMC.

Two important special cases of WMC are \emph{weighted \sCNF} and \emph{weighted \sDNF}, which require the input formula to be in conjunctive normal form (\CNF) and disjunctive normal form (\DNF), respectively. Inference in probabilistic graphical models typically reduces to solving weighted \sCNF instances, while query evaluation in probabilistic databases reduces to solving weighted \sDNF instances~\cite{Suciu-PDBs}. However, both weighted \sCNF and weighted \sDNF are known to be \sharpP-hard~\cite{Valiant79}, and this computational complexity is a major bottleneck for solving large-scale \WMC instances.

To overcome this problem, two main paradigms have been developed. The first paradigm is \emph{knowledge~compilation}~\cite{Cadoli97,SeKa96}, which solves computationally difficult problems by compiling them into a new representation (i.e., a target language), where they can be subsequently solved efficiently. Following compilation, exact inference in \WMC can be done in linear time~\cite{DaMA11}. However, the compilation process can produce exponentially-sized problem representations~(i.e., arithmetic circuits). Furthermore, knowledge compilation is not robust to changes:  for every change in the underlying model, the computationally demanding knowledge compilation process needs to be repeated. As a result, approaches based on knowledge compilation struggle to scale to large and varying problem instances.  

The second paradigm is approximate solving~\cite{Ermon13,CMV16,Meel17-DNF}, which provides approximations of the model count as opposed to an exact solution.  Loosening the requirement for exactitude renders \WMC more tractable, especially for \sDNF counting, where approximate solving admits a fully polynomial randomized approximation scheme (FPRAS) due to Karp, Luby, and Madras~\shortcite{Karp89}, which we denote KLM. KLM allows for faster estimation of \sDNF model counts, while also providing probabilistic guarantees on its approximations, and it is the state of the art for weighted \sDNF. Nonetheless, KLM runs in $O(nm)$, where $n$ denotes the number of variables and $m$ the number of clauses of the input DNF formula. Hence, KLM struggles to scale to real-world \DNF formulas. 

In this work, we propose Neural\sDNF, an approach that combines deep learning and approximate model counting and enables fast weighted \sDNF approximation. We first generate instances of weighted \sDNF and solve them using KLM to produce training data. We then use a graph neural network (GNN) to capture the symbolic structure of DNF formulas and train our system. By construction, Neural\sDNF produces approximations in $O(m\bar{w})$, where $\bar{w}$ denotes the average clause width. This reduces to just $O(n+m)$ for bounded width. Our approach does not provide guarantees as with KLM, but instead enables a relative speed-up of multiple orders of magnitude in the average case. This is especially true, since, in practice, $\bar{w} << n$.  
Our experiments show that the GNN learns to accurately estimate weighted model counts and generalizes to novel formulas. Indeed, our model computes solutions to unseen weighted \sDNF instances with $99\%$ accuracy relative to an additive error threshold of $0.1$ with respect to tight KLM approximations. It also generalizes to larger problem instances involving up to 15K variables remarkably well, despite only seeing formulas with at most 5K variables during training.

In summary, Neural\sDNF makes the following contributions:
\begin{itemize}
\item It produces \emph{efficient} and \emph{highly accurate} weighted \sDNF approximations in $O(m\bar{w})$, and in \emph{linear time} with boun\-ded width. 
\item It \emph{reliably scales} to \sDNF instances with up to 15K variables, which, to our knowledge, is a first for neural-sym\-bolic methods. 
\item It is \emph{robust} in that it can produce approximations for any problem instance over a given domain following training.
\end{itemize}
Our findings suggest that GNNs can effectively and efficiently perform large-scale \sDNF through training with dense and reliable data. Further experiments and details are deferred to the appendix of this paper.\footnote{The extended version of this paper including the appendix is available at:  arxiv.org/pdf/1904.02688.pdf.}

 \section{Preliminaries}
We briefly introduce weighted model counting, the KLM algorithm, and graph neural networks.

\subsection{Weighted Model Counting}
\label{ssec:wmc}
Given a (finite) set $S$ of propositional variables, a \emph{literal} is of the form $v$, or $\neg v$, where $v \in S$. A \emph{conjunctive clause} is a conjunction of literals, and a \emph{disjunctive clause} is a disjunction of literals. A clause has \emph{width} $k$ if it has exactly $k$ literals. A formula $\phi$ is in \emph{conjunctive normal form (\CNF)} if it is a conjunction of disjunctive clauses, and it is in \emph{disjunctive normal form (\DNF)} if it is a disjunction of conjunctive clauses. We say that a \DNF (resp., \CNF) has width $k$ if it contains clauses of width at most $k$.  %To illustrate, the formula ${\phi=(x_1 \land \neg x_3) \vee (x_4 \land x_1)}$ is a \DNF with clauses of width 2.
An assignment $\nu: S \mapsto \{0,1\}$ maps every variable to either $0$ (false), or $1$ (true). An assignment $\nu$ \emph{satisfies} a propositional formula $\phi$, denoted $\nu\models \phi$, in the usual sense, where $\models$ is the propositional entailment relation.

Given a propositional formula $\phi$, its \emph{model count} $\#\phi$ is the number of assignments $\nu$ satisfying $\phi$. The \emph{weighted model count} of $\phi$ is given by ${\sum_{\nu \models \phi} w(\nu)}$, where  $w: \mathfrak{A}\mapsto \mathbb{R}$ is a \emph{weight function}, and $\mathfrak{A}$ is the set of all possible assignments. In this work, we set $w: \mathfrak{A} \mapsto [0,1] \cap \mathbb{Q}$ such that every assignment is mapped to a rational probability and ${\sum_{\nu \in \mathfrak{A}} w(\nu) = 1}$. As common in the literature, we view every propositional variable as an independent Bernoulli random variable and assign probabilities to literals. 

\subsection{The KLM Algorithm} 
Exactly solving weighted \sDNF instances is \sharpP-hard and thus intractable. The KLM algorithm~\cite{Karp89} is a fully polynomial randomized approximation scheme (FPRAS), and provides probabilistic guarantees for weighted \sDNF. More formally, given an error $\epsilon>0$ and a confidence value ${0< \delta < 1}$, KLM computes $\hat{\mu}$, an approximation of the true weighted model count $\mu$, in polynomial time such that %
$
{\Pr\big(\mu (1 - \epsilon) \leq \hat{\mu} \leq \mu (1 + \epsilon)\big) \geq 1 - \delta}.
$

Specifically, for a \DNF $\phi$ with $n$ variables and $m$ clauses, KLM computes a number of sampling trials ${\tau = 8(1+\epsilon) m \log({\frac{2}{\delta}})\frac{1}{\epsilon^2}}$, and initializes a trial counter $N$ to~$0$.  Then, at every trial, KLM performs the following steps:
\begin{enumerate}
\item If no current sample assignment exists, randomly select a clause $C_i$ with probability $\frac{p(C_i)}{\sum_{j=1}^m p(C_j)}$, then randomly generate a satisfying assignment for $C_i$ using the variable probability distribution.
\item Check whether the current assignment satisfies a randomly selected clause $C_k$. If so, increment $N$ and generate a new sample assignment. Otherwise, do nothing.
\end{enumerate}
KLM  returns $\frac{\tau\sum_{j=1}^m p(C_j)}{mN}$ as an estimate for the weighted DNF count. Since assignment checking runs in $O(n)$, the complexity of KLM amounts to $O\big(nm \epsilon^{-2}\log(\frac{1}{\delta})\big)$. KLM is the state of the art for approximate weighted \sDNF, so we use it to label \DNF formulas used to train our model. %Further details about KLM can be found in \cite{Karp89,Karp-ASFCS1983}

\begin{figure*}[t]  % So that it spans both columns
	\centering
	\begin{subfigure}[t]{1.3in}
		\begin{tikzpicture}[node distance = 1cm,line width=0.8pt,shorten >=2pt, shorten <=2pt,->]
		\tikzstyle{var} = [text width=1.6em, text centered,text=white,fill=gray!120, circle,inner sep=2pt]
		\node[var] (x1) {$x_1$};
		\node[var,below=0.15 of x1] (x1n) {$\neg x_1$};
		\node[var,below=0.15 of x1n] (x2) {$x_2$};
		\node[var,below=0.15 of x2] (x2n) {$\neg x_2$};
		\node[var,below right=0.3cm and 0.6cm of x1] (c1) {$c_1$};
		\node[var,below=0.3cm of c1] (c2) {$c_2$};
		\node[var,text=black,opacity=0.5, below right=0cm and 0.6cm of c1] (d) {$d$};
		\path[draw] (x1) to (c1);
		\path[draw] (x2) to(c1);
		\path[draw] (x1n) to (c2);
		\path[draw] (x2n) to(c2);
		\end{tikzpicture}
		\caption{}\label{fig:1a}		
	\end{subfigure}
	\hfill
	\begin{subfigure}[t]{1.3in}
		\begin{tikzpicture}[node distance = 1cm,line width=0.8pt,shorten >=2pt, shorten <=2pt,->]
		\tikzstyle{var} = [text width=1.6em, text centered,,text=white, fill=gray!120, circle,inner sep=2pt]
		\node[var,text=black,opacity=0.5] (x1) {$x_1$};
		\node[var,text=black,opacity=0.5, below=0.15 of x1] (x1n) {$\neg x_1$};
		\node[var, text=black,opacity=0.5,below=0.15 of x1n] (x2) {$x_2$};
		\node[var,text=black,opacity=0.5, below=0.15 of x2] (x2n) {$\neg x_2$};
		\node[var,below right=0.3cm and 0.6cm  of x1] (c1) {$c_1$};
		\node[var,below=0.3cm of c1] (c2) {$c_2$};
		\node[var,below right=0cm and 0.6cm  of c1] (d) {$d$};
		\path[draw, very thick, gray!100] (c1) to (d);
		\path[draw,very thick, gray!100] (c2) to (d);
		\end{tikzpicture}
		\caption{}
		\label{fig:1b}		
	\end{subfigure}
	\hfill
	\begin{subfigure}[t]{1.3in}
		\begin{tikzpicture}[node distance = 1cm,line width=0.8pt,shorten >=2pt, shorten <=2pt,->]
		\tikzstyle{var} = [text width=1.6em, text centered,,text=white, fill=gray!120, circle,inner sep=2pt]
		\node[var, text=black,opacity=0.5] (x1) {$x_1$};
		\node[var,text=black,opacity=0.5, below=0.15 of x1] (x1n) {$\neg x_1$};
		\node[var,text=black,opacity=0.5,below=0.15 of x1n] (x2) {$x_2$};
		\node[var,text=black,opacity=0.5,below=0.15 of x2] (x2n) {$\neg x_2$};
		\node[var,below right=0.3cm and 0.6cm  of x1] (c1) {$c_1$};
		\node[var,below=0.3cm of c1] (c2) {$c_2$};
		\node[var,below right=0cm and 0.6cm  of c1] (d) {$d$};
		\path[draw,very thick, gray!100] (d) to (c1);
		\path[draw,very thick, gray!100] (d) to (c2);
		\end{tikzpicture}
		\caption{}
		\label{fig:1c}		
	\end{subfigure}
	\hfill
	\begin{subfigure}[t]{1.3in}
		\begin{tikzpicture}[node distance = 1cm,line width=0.8pt,shorten >=2pt, shorten <=2pt,->]
		\tikzstyle{var} = [text width=1.6em, text centered,text=white, fill=gray!120, circle,inner sep=2pt]
		\node[var] (x1) {$x_1$};
		\node[var,below=0.15 of x1] (x1n) {$\neg x_1$};
		\node[var,below=0.15 of x1n] (x2) {$x_2$};
		\node[var,below=0.15 of x2] (x2n) {$\neg x_2$};
		\node[var,below right=0.3cm and 0.6cm  of x1] (c1) {$c_1$};
		\node[var,below=0.3cm of c1] (c2) {$c_2$};
		\node[var,opacity=0.5,text=black,below right=0cm and 0.6cm  of c1] (d) {$d$};
		\path[draw, dashed] (x1) to[out=-145,in=130]  (x1n);
		\path[draw, dashed] (x1n) to[out=30,in=-55]  (x1);
		\path[draw, dashed] (x2n) to[out=145,in=-130] (x2);
		\path[draw, dashed] (x2) to[out=-30,in=55]  (x2n);
		\path[draw, thick] (c1) to (x1);
		\path[draw,thick] (c1) to (x2);
		\path[draw,thick] (c2) to (x1n);
		\path[draw,thick] (c2) to (x2n);
		\end{tikzpicture} 
		\caption{}
		\label{fig:1d}		
	\end{subfigure}
	\hfill
	\caption{Message passing protocol on the DNF formula $\psi=(x_1 \land x_2) \vee (\neg x_1 \land \neg x_2)$.}
	\label{fig:OverallDiagram}
\end{figure*}
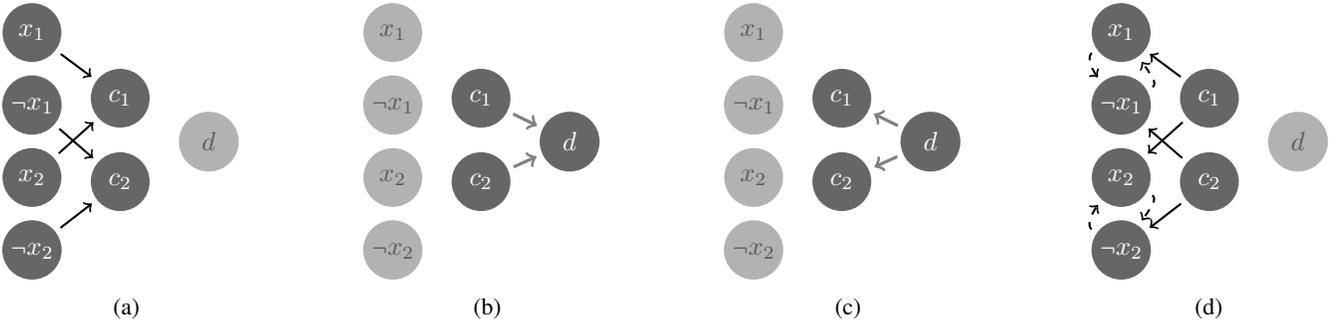

\subsection{Graph Neural Networks}

Graph neural networks (GNNs) \cite{Gori2005,Scarselli09} are neural networks specifically designed to process structured graph data. 
In a GNN, every graph node $x$ is given a vector representation $v_{x}$, which is updated iteratively. A node $x$ receives information from its \emph{neighborhood} $N(x)$, which is the set of nodes connected by an edge to $x$. 
Let $v_{x,t}$ denote the value of $v_{x}$ at iteration $t$. We write a node update as: 
\[
v_{x,t+1} = combine \Big(v_{x,t},aggregate\big(N(x)\big)\Big),
\] 
where \emph{combine} and \emph{aggregate} are functions, and \emph{aggregate} is permutation-invariant. 
We use layer-norm LSTMs \cite{Ba16} as our \emph{combine} function, and sum as our \emph{aggregate} function. This is similar to gated graph neural networks \cite{Li15}, except that we replace the gated recurrent unit (GRU) \cite{Chung14} with a layer-norm LSTM, given the remarkable empirical success of the latter~\cite{Selsam-ICLR2019,Prates-AAAI2019}. Upon termination of all iterations, the final node representations are used to compute the target output. 

GNNs are highly expressive computational models: GNNs can be as discerning between graphs as the Weisfeiler-Lehman (WL) graph isomorphism heuristic \cite{Keyulu18,GroheAAAI19}.  Unlike feature engineering \cite{Kashima03} and static embedding methods \cite{Wang14}, GNNs can autonomously learn relationships between nodes, identify important features, and build models that can generalize to unseen graphs. 

\section{Graph Neural Network Model}
\label{sec:GNN}
We propose Neural\sDNF, a new method for solving weighted \#DNF problems based on GNNs. We model DNF formulas as graphs, and use a GNN to iterate over these graphs to compute an approximate weighted model count. 

\subsection{Model Setup}
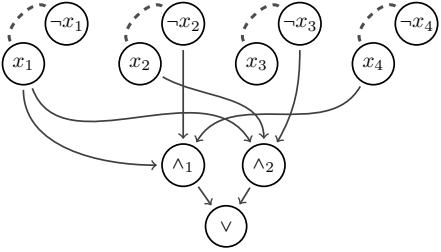
\begin{figure}[t]
	\centering
	\scalebox{0.81}{	
		\begin{tikzpicture}[node distance = 1cm,line width=0.8pt,shorten >=2pt, shorten <=2pt,->]
		\tikzstyle{var} = [text width=1.6em, text centered, circle, draw,inner sep=1pt]
		\node[var] (x1) {$x_1$};
		\node[var,above right=0.15cm and 0.2cm of x1](x1n) {$\neg x_1$};
		\node[var,right=1.2cm of x1] (x2){$x_2$};
		\node[var,above right=0.15cm and 0.2cm of x2](x2n) {$\neg x_2$};
		\node[var,right=1.2cm of x2](x3) {$x_3$};
		\node[var,above right=0.15cm and 0.2cm of x3](x3n) {$\neg x_3$};
		\node[var,right=1.2cm of x3] (x4){$x_4$};
		\node[var,above right=0.15cm and 0.2cm of x4](x4n) {$\neg x_4$};
		\node[var,below =1.6cm of x2n] (c1) {$\land_1$};
		\node[var,right=0.6cm of c1] (c2) {$\land_2$};
		\node[var,below right=0.5cm and 0.2cm  of c1] (d) {$\vee$};
		%
		%\draw[blue,dash pattern= on 3pt off 5pt] (x1) to[out=90,in=180]  (x1n);
		\draw[black!70,line width=0.5mm, dashed,-] (x1) to[out=120,in=140]   (x1n);
		\draw[black!70,line width=0.5mm,dashed, -]  (x2) to[out=120,in=140](x2n);
		\draw[black!70,line width=0.5mm,dashed, -]  (x3) to[out=120,in=140](x3n);
		\draw[black!70,line width=0.5mm, dashed,-]  (x4) to[out=120,in=140] (x4n);
		\path[draw, gray!150] (x1) to[out=-90,in=180]  (c1);
		\path[draw, gray!150] (x2n) to[out=-90,in=90]  (c1);
		\path[draw, gray!150] (x4) to[out=-120,in=60]  (c1);
		\path[draw,gray!150] (x1) to[out=-70,in=120]  (c2);
		\path[draw,gray!150] (x2) to[out=-30,in=90]   (c2);
		\path[draw,gray!150] (x3n) to[out=-90,in=60]  (c2);
		\path[draw,gray!150] (c1)  to (d);
		\path[draw,gray!150] (c2) to (d);
		\end{tikzpicture}
	}	
	\caption{Graph encoding of the DNF formula {${\phi= (x_1 \land \neg x_2 \land x_4) \vee (x_1 \land x_2 \land \neg x_3)}$}.}
	\label{fig:DNFGraph}
\end{figure}

We encode a DNF formula as a graph with 3 layers as shown in Figure~\ref{fig:DNFGraph}: a \emph{literal} layer, a \emph{conjunction} layer, and a \emph{disjunction} layer. In the \emph{literal} layer, every DNF variable is represented by 2 nodes corresponding to its positive and negative literals, which are connected by a (dashed) edge to highlight that they are complementary. In the \emph{conjunction} layer, every node represents a conjunction and is connected to \emph{literal} nodes whose literals appear in the conjunction.  Finally, the \emph{disjunction} layer contains a single disjunction node, which is connected to all nodes in the \emph{conjunction} layer.

To approximate the model count of a \DNF formula, we use a message-passing GNN model that iterates over the corresponding \DNF graph and returns a Gaussian distribution. Initially, the network computes vector representations for all literal nodes, given their probabilities, using a multi-layer perceptron (MLP) $f_{enc}$. More formally, a $k-$dimensional representation $v_{{x_i},0}$ of a literal $x_i$ with probability $p_i$ is computed as $v_{{x_i},0} = f_{enc}(p_i)$.
Nodes in the \emph{conjunction} and \emph{disjunction} layers are initialized to two representation vectors $v_c$ and $v_d$, respectively, and the values for these vectors are learned over the course of training. After initialization, node representations are updated across~$T$ message passing iterations.
\subsection{Message Passing Protocol}
A message passing iteration consists of the following 4 steps:

\noindent
\textbf{(a)} \emph{Literal} layer nodes compute messages using an MLP $M_l$ and pass them to their neighboring \emph{conjunction} layer nodes. These \emph{conjunction} nodes then aggregate these messages using the sum function and update their representation using a layer-norm LSTM $L_{c_1}$. The updated \emph{conjunction} node representations, denoted $\hat{v}_{x_c,t+1}$, are given formally as:
\begin{align*}
\hat{v}_{{x_{c}},t+1} = L_{c_1}\Big(v_{{x_{c}},t}, \sum_{x_l \in N(x_{l})}M_l(v_{x_l,t})\Big).
\end{align*}

\noindent
\textbf{(b)}  \emph{Conjunction} layer nodes compute and send messages to the disjunction node via an MLP $M_c$. The disjunction node aggregates these and updates using a layer-norm LSTM $L_d$, i.e.,
\begin{align*}
	v_{{x_{d}},t+1} = L_{d}\Big(v_{{x_{d}},t}, \sum_{x_c \in N(x_{d})}M_c(\hat{v}_{x_c,t+1})\Big).
\end{align*}

\noindent
\textbf{(c)}  The disjunction node computes a message using an MLP $M_d$ and sends it to the \emph{conjunction} nodes, which update their representation using a different LSTM cell $L_{c_2}$: 
\begin{align*}
 v_{{x_{c}},t+1} = L_{c_2}\Big(\hat{v}_{{x_{c}},t+1}, M_d(v_{x_d,t+1})\Big).
\end{align*}

\noindent
\textbf{(d)} Using their latest representations, \emph{conjunction} nodes send messages to neighboring nodes in the \emph{literal} layer. \emph{Literal} layer nodes aggregate these messages and concatenate them (represented with $||$) with messages from their corresponding negated literal. Then, they use this message to update their representations using a layer-norm LSTM $L_l$:
\begin{align*}
v_{{x_{l}},t+1} &= L_l\Big(v_{{x_{l}},t},\big(\sum_{x_c \in N(x_{l})} M_c(v_{{x_c},t+1}) || M_l(v_{{\neg x_{l}},t})\big)\Big).
\end{align*}

A visual representation of the 4 message passing steps for a simple formula  is provided in Figure~\ref{fig:OverallDiagram}. In this protocol, we use 2 distinct LSTM cells $L_{c_1}$ and $L_{c_2}$ to update the representations of \emph{conjunction} nodes at steps (a) and (c), so that the network learns separate update procedures for literal-based and disjunction-based updates. 
%
%The 4 steps of a message passing iteration can be mathematically formulated  as follows:
%\begin{align*}
%\text{a.}\ \hat{v}_{{x_{c}},t+1} &= L_{c1}\Big(v_{{x_{c}},t}, \sum_{x_l \in N(x_{l})}M_l(v_{x_l,t})\Big)\\
%\text{b.}\  v_{{x_{d}},t+1} &= L_{d}\Big(v_{{x_{d}},t}, \sum_{x_c \in N(x_{d})}M_c(\hat{v}_{x_c,t+1})\Big)\\
%\text{c.}\  v_{{x_{c}},t+1} &= L_{c2}\Big(\hat{v}_{{x_{c}},t+1}, M_d(v_{x_d,t+1})\Big)\\
%\text{d.}\  v_{{x_{l}},t+1} &= L_l\Big(v_{{x_{l}},t},\big(\sum_{x_c \in N(x_{l})} M_c(v_{x,t+1}) || M_l(v_{\overline{x_{l}},t})\big)\Big),
%\end{align*} 
%where $||$ denotes the concatenation operation. A visual representation of the 4 steps of the message passing protocol on a simple formula is provided in Fig. \ref{fig:OverallDiagram}
%
At the end of message passing, the final disjunction node representation $v_{x_d,T}$ is passed through an MLP $f_{out}$. The final layer of this MLP consists of two neurons $n_\mu$ and $n_\sigma$, which  return the mean and standard deviation, respectively, of a predicted Gaussian distribution. 
\subsection{Loss Function}
\label{ssec:loss}
Given $\epsilon$ and $\delta$, KLM returns an estimate $\hat{\mu}$ of the true model count $\mu$ within a multiplicative bound with respect to $\epsilon$, and this bound holds with probability $1 - \delta$. By identifying different configurations of $\epsilon$ and $\delta$ that lead to an identical KLM running time, one can deduce that the probability mass is concentrated around $\hat{\mu}$ and decays away from it, and this holds for all DNFs. Note that the multiplicity of the bound interval on $\hat{\mu}$ w.r.t. $\epsilon$ makes it hard to fit standard distributions on it. Hence, we apply a natural logarithm to this bound to get the additive bound on $\log{\mu}$: 
\[
\log{\hat{\mu}} - \log{(1 + \epsilon)} \leq \log{\mu} \leq \log{\hat{\mu}} + \log{(1 + \epsilon)}.
\]
We can then fit a Gaussian $\mathcal{N}(\mu',\sigma)$ to this bound by setting  $\mu'=\hat{\mu}\  \text{and}\  \sigma=\sfrac{\log{(1+\epsilon)}}{F^{-1}(1 - \frac{\delta}{2})}$, where $F^{-1}$ denotes the inverse cumulative distribution function of the standard Gaussian distribution. The GNN is thus trained to predict $\log{\mu}$, a negative number. We adapt the exponential linear unit (ELU) \cite{Djork16} activation function and apply it to $n_\mu$ and $n_\sigma$.
More specifically, we use 
\[ELU+1(x) = \begin{cases} 
e^{-x} & \text{if} \, x \leq 0 \\ x + 1 & \text{otherwise,} \end{cases}
\] 
such that $n_{\mu}$ uses $-ELU+1(x)$, and $n_{\sigma}$ uses $ELU+1(x)$, thereby restricting their outputs to be negative and positive, respectively. 

To compare the predicted Gaussian and the KLM result, we use Kullback-Leibler~(KL) divergence, which for two Gaussians $\mathcal{N}_1(\mu_1, \sigma_1)$ and $\mathcal{N}_2(\mu_2, \sigma_2)$ is given by: 
\[
KL(\mathcal{N}_1,\mathcal{N}_2) = \log{\frac{\sigma_2}{\sigma_1}} - \frac{1}{2} + \frac{\sigma_{1}^{2} + (\mu_{1} - \mu_2)^2}{2 \sigma_{2}^{2}}.
\]
We set $\mathcal{N}_1$ to be the prediction returned by the network and $\mathcal{N}_2$ to be the KLM approximation. This choice is critical in order to avoid the system minimizing the training loss by learning to produce arbitrarily large values of $\sigma_2$.

\begin{table}[t]
	\centering
	\caption{Distribution of formula sizes in the training set.}
	\begin{tabular}{lcccc}
		\toprule
		\text{Size ($n$)} & 50 & 100 & 250 & 500\\
		\cmidrule(l){2-5}	
		\text{Count} & 30000 & 20000 & 16000 & 12000\\
		\midrule
				\midrule
		\text{Size ($n$)} & 750 & 1000 & 2500 & 5000\\
		\cmidrule(l){2-5}	
		\text{Count} & 10000 & 8000 & 6000  & 3000\\
		\bottomrule
	\end{tabular}
	\label{tab:varSizes}
\end{table}

\section{Experiments}
\label{sec:exp}
We train our model on a large set of DNF formulas and measure its generalization relative to new DNF formulas. These formulas are distinct in terms of \emph{structure} (i.e., the underlying clauses and variables in every clause) and \emph{size} (i.e., the number of clauses and variables is larger), so our experiments target generalization in both aspects. To evaluate \emph{structure generalization}, we run our GNN on unseen formulas of comparable size to training formulas and measure its performance. To evaluate \emph{size generalization}, we run tests on novel, larger formulas and assess how well the GNN performs.
To further validate our model and data generation procedure, we also run these experiments using differently generated synthetic datasets \cite{Meel18}. 

\subsection{Experimental Setup}
\label{ssec:expSet}
In our experiments, we compare Neural\sDNF predictions $\hat{\mu}$ with those of KLM and check whether their absolute difference falls within pre-defined additive thresholds. We opt for additive error, as opposed to multiplicative error, as the former produces an absolute distance metric, whereas the latter is relative to the model count.

Owing to the lack of standardized benchmarks, we generate synthetic formulas using a novel randomized procedure designed to produce more variable formulas. We generate 100K distinct training formulas, where formula counts per $n$ are shown in Table \ref{tab:varSizes}. For every $n$, formulas are generated with fixed clause width $w \in\{3, 5, 8, 13, 21, 34\}$ and number of clauses $m$ from $\{0.25, 0.375, 0.5, 0.625, 0.75\}\cdot n$, such that every valid setting (i.e., all configurations except $w=3$ and $m=0.25\cdot n$) is represented equally, and each formula has 4 variable probability distributions. More details about our data generation can be found in the appendix.
\begin{figure}[t]
	\centering
	\begin{tikzpicture}	[scale=0.85]
	\begin{axis}[
	width=0.5\textwidth,
	tick align=outside,
	tick pos=left,
	ylabel near ticks,
	xlabel near ticks,
	x grid style={white!69.01960784313725!black},
	xlabel={KLM Approximations},
	xmin=-0.5, xmax=600,
	xtick={0,60,120,180,240,300,360,420,480,540,600},
	xticklabels={0.0,0.1,0.2,0.3,0.4,0.5,0.6,0.7,0.8,0.9,1.0},
	y grid style={white!69.01960784313725!black},
	ylabel={GNN Approximations},
	ymin=-0.5, ymax=600,
	ytick={0,60,120,180,240,300,360,420,480,540,600},
	yticklabels={0.0,0.1,0.2,0.3,0.4,0.5,0.6,0.7,0.8,0.9,1.0}
	]
	\addplot graphics [includegraphics cmd=\pgfimage,xmin=-0.5, xmax=599.5, ymin=599.5, ymax=-0.5] {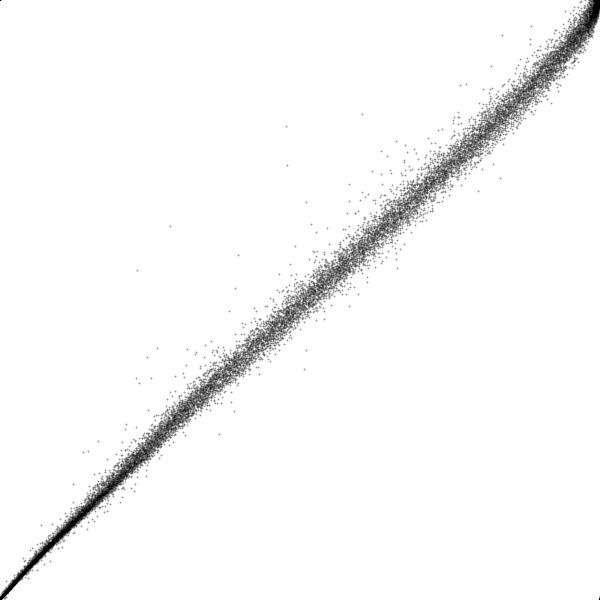};
	\end{axis}
	\end{tikzpicture}
	\caption{A gray-scale heat map representing the distribution of GNN predictions compared to KLM approximations.}
	\label{fig:heatmap}	
\end{figure}
The \emph{structure} evaluation test set is generated analogously, and contains 13080 distinct formulas. The \emph{size} evaluation set contains 348 formulas with $n=10K$ and 116 formulas with $n=15K$, with one probability distribution each.

For all experiments, we use $k=128$-dimensional vector representations. We define $f_{enc}$ as a 3-layer MLP with layer sizes 8, 32, and 128, message-generating MLPs ($M_l$, $M_c$, and $M_d$) as 4-layer MLPs with 128-sized layers, and $f_{out}$ as a 3-layer MLP with layers of size 32, 8, and 2. We use the rectified linear unit (ReLU) as the activation function at MLP hidden layers, and linear activation at the output layer for all MLPs except $f_{out}$. For $f_{out}$, output activation is as defined in Section \ref{ssec:loss}. Generated formulas were labelled using KLM with $\epsilon=0.1$ and $\delta=0.05$ to achieve a reasonable trade-off between label accuracy and generation tractability.

We train the system for 4 epochs on a P100 GPU using KL divergence loss, the Adam optimizer \cite{Kingma-ICLR2014}, a learning rate of $\lambda = 10^{-5}$, a gradient clipping ratio of 0.5, and ${T=8}$ message passing iterations.

\subsection{Results}
\begin{table}[t] 
	\centering
	\caption{GNN accuracy (\%) w.r.t.\ to additive thresholds.} 
	\label{tab:testSet} 
	\begin{tabular}{lcccc}
		\toprule 
		\multirow{2}{*}{\textbf{Evaluation Data}} & \multicolumn{4}{c}{\textbf{Thresholds}} \\
		\cmidrule(r){2-5}
		& 0.02 & 0.05 & 0.10 & 0.15  \\
		\midrule 
		\text{Training Set} & 87.14 & 98.80 & 99.97 & 99.99 \\
		\text{Test Set} & 87.37 & 98.76 & 99.95 & 99.98 \\
		\bottomrule
	\end{tabular}
\end{table}
On the \emph{structure} generalization test, Neural\sDNF predictions align very closely with those of KLM, as shown in Figure~\ref{fig:heatmap}. 
The model is within $0.02$ of the KLM \WMC estimate over $87.37\%$ of the test set, and this rises to 99.95\% for a threshold of $0.1$. The model also performs consistently across different $n$, with accuracy varying by at most $4.5\%$ between any two different $n$ values for all four test thresholds. Overall test results are given in Table~\ref{tab:testSet}. 

The proximity between training and testing accuracies at all thresholds shows that the network has not fit or memorized its training formulas, but has instead learned a general \WMC procedure. The results parametrized by $n$ are provided in Table~\ref{tab:byNV}. 
These results show that the network maintains a high accuracy (e.g., $95.5\%$ for threshold 0.05) across all $n$ values, and so does not rely on a particular $n$ to achieve its high overall performance. Notably, the model is also robust against variation in $w$. As shown in Table~\ref{tab:byCW}, the model scores above $96\%$ and $99\%$ across all widths for thresholds $0.05$ and $0.1$, respectively. Interestingly, it has near-perfect performance for larger widths 13, 21, and 34, where weighted model counts are near-zero, and has relatively higher accuracy at threshold $0.02$ when $w=3$, where counts are almost one. Simultaneously performing well in both extreme cases, coupled with high accuracy on intermediate widths, further highlights the robustness of our model. 

On the \emph{size} generalization task, our model maintains accuracies of $97.13\%$ and $94.83\%$ with a threshold of $0.1$ on 10K and 15K-variable formulas, respectively, despite having as many as triple the variables as in training. The full results for \emph{size} generalization are given in Table~\ref{tab:sizeGen}. The same results parametrized by width $w$ are also given in Table~\ref{tab:sizeGenbyCW}. These show that the network performs consistently across widths 3, 5, and 8, but performs less well at $w=13$. This is due to formulas with $w=13$ exhibiting a ``phase transition'' at this $n$ and $m$. Indeed, in this setting, model counts fluctuate dramatically, since $m$ is in the same order of magnitude as $p^{-1}$, the inverse expected clause satisfaction probability. In the training set, this phenomenon occurs at smaller widths, but never for $w=13$, so this is an entirely new situation for the model, at a much larger scale. Nonetheless, it achieves an encouraging accuracy of $82.5\%$ for the threshold $0.1$. 

\begin{table}[t]
	\centering
	\caption{GNN accuracy (\%) over test set by threshold versus number of formula variables ($n$).}
	\label{tab:byNV} 
	\begin{tabular}{lcccc}
		\toprule % <-- Toprule here
		\multirow{2}{*}{\textbf{ $n$}} & \multicolumn{4}{c}{\textbf{Thresholds}}\\
		\cmidrule(l){2-5}
		& 0.02 & 0.05 & 0.10 & 0.15\\
		\midrule % <-- Midrule here
		\textbf{50} & 85.58 & 98.58 & 99.98 & 100.0\\
		\textbf{100} & 87.87 & 98.87 & 100.0 & 100.0\\
		\textbf{250} & 87.93 & 99.24 & 100.0 & 100.0\\
		\textbf{500} & 87.67 & 99.40 & 99.99 & 100.0\\
		\textbf{750} & 87.56 & 99.15 & 100.0 & 100.0\\
		\textbf{1000} & 86.79 & 99.01 & 99.98 & 100.0\\
		\textbf{2500} & 90.06 & 98.17 & 99.85 & 99.94\\
		\textbf{5000} & 88.15 & 95.86 & 99.48 & 99.74\\			
		\bottomrule
	\end{tabular}
\end{table}
\begin{table}[t]
	\centering
	\caption{GNN accuracy (\%) over test set by threshold versus formula clause widths ($w$).}
	\label{tab:byCW} 
	\begin{tabular}{lcccc}
		\toprule 
		\multirow{2}{*}{$w$} & \multicolumn{4}{c}{\textbf{Thresholds}}\\
		\cmidrule(l){2-5}
		& 0.02 & 0.05 & 0.10 & 0.15\\
		\midrule % <-- Midrule here
		\textbf{3} & 80.42 & 98.66 & 99.87 & 99.93\\
		\textbf{5} & 68.04 & 96.56 & 99.90 & 99.98\\
		\textbf{8} & 79.10 & 97.77 & 99.96 & 99.98\\
		\textbf{13} & 99.70 & 99.98 & 100.0 & 100.0\\
		\textbf{21} & 100.0 & 100.0 & 100.0 & 100.0\\	
		\textbf{34} & 100.0 & 100.0 & 100.0 & 100.0\\			
		\bottomrule
	\end{tabular}
\end{table}

These results show that reliable approximate model counting on large-scale formulas can be achieved, even with training restricted to smaller formulas. From a practical perspective, this gives further evidence that large-scale solvers can be trained using smaller formulas that are tractably labelled with existing solvers. In additional experiments run on differently generated datasets (cf. appendix), our system also maintains very high performance, and in fact performs better on fully random formulas \cite{Meel18} than on formulas generated with our protocol. This further highlights the robustness of Neural\sDNF, and validates the quality of our data generation procedure.

For all these results, message passing iterations are essential. Indeed, when run with just 2 message passing iterations for our ablation study, the GNN performs significantly worse across all experiments. However, this does not imply that performance always improves with more message passing. In fact, running too many message passing iterations makes the system prone to overfitting: When run with 32 iterations, the system achieves a similar performance in structure generalization, but its performance drops significantly in size generalization. This shows that a trade-off value of message passing iterations, in our case 8, must be selected, to enable sufficient communication, while not encouraging overfitting. Further details on our ablation study and experiments with 32 iterations can be found in the appendix.

All in all, our model achieves remarkable performance both in terms of structure and size generalization. These results highlight the power and scalability of neural message passing (NMP) methods to perform advanced reasoning tasks, and therefore justify further consideration of NMP. %for approximation tasks in reasoning.

\begin{table}[t]
	\centering
	\caption{Accuracy (\%) by threshold with respect to additive thresholds on \emph{size} generalization test formulas.} 
	\begin{tabular}{lcccc}
		\toprule % <-- Toprule here
		\multirow{2}{*}{\textbf{$n$}} & \multicolumn{4}{c}{\textbf{Thresholds}} \\
		\cmidrule(l){2-5}
		& 0.02 & 0.05 & 0.10 & 0.15  \\
		\midrule % <-- Midrule here
		\textbf{10K} & 79.89 & 89.94 & 97.13 & 99.71 \\
		\textbf{15K} & 72.41 & 81.90 & 94.83 & 97.41 \\
		\bottomrule
	\end{tabular}
	\label{tab:sizeGen} 
	\hfill
\end{table}
\begin{table}
	\centering
	\caption{Accuracy (\%) by threshold over \emph{size} generalization test formulas versus $w$.}
	\label{tab:sizeGenbyCW} 
	\begin{tabular}{lcccc}
		\toprule 
		\multirow{2}{*}{$w$} & \multicolumn{4}{c}{\textbf{Thresholds}}\\
		\cmidrule(l){2-5}
		& 0.02 & 0.05 & 0.10 & 0.15\\
		\midrule % <-- Midrule here
		\textbf{3} & 78.13 & 90.63 & 98.44 & 100.0\\
		\textbf{5} & 73.75 & 90.0 & 100.0 & 100.0\\
		\textbf{8} & 76.25 & 91.25 & 98.75 & 100.0\\
		\textbf{13} & 40.0 & 56.25 & 82.5 & 95.0\\
		\textbf{21,34} & 100.0 & 100.0 & 100.0 & 100.0\\	
		%	\textbf{34} & 100.0 & 100.0 & 100.0 & 100.0\\			
		\bottomrule
	\end{tabular}
\end{table}

\subsection{Running Time Analysis}
In the average case, our GNN runs in $O(m\bar{w})$, where $\bar{w}$ denotes the average formula clause width. By contrast, KLM runs in $O(nm)$, so is much slower for standard cases in practice, where $\bar{w} << n$. 
In the worst case, our GNN runs in $O(nm)$, which is asymptotically identical to KLM. However, in a best-case scenario where $\bar{w}$ is upper-bounded, the GNN complexity drops to just $O(n+m)$, enabling linear-time approximations to be made, whereas KLM remains $O(nm)$, since its complexity does not depend on $\bar{w}$. % Not sure about this comment.
Hence, our system enables much faster approximations than KLM in practice, where $\bar{w} << n$, and these approximations are in linear time with bounded clause width $\bar{w}$. A detailed complexity analysis for Neural\sDNF can be found in the appendix.

\begin{table}[t]
	\centering
	\caption{Runtimes (s) for KLM and our GNN by number of variables ($n$),  with $w=3, 34$ and $m$ = 0.75$n$.}
	\label{tab:runtimes}
	\begin{tabular}{lccccc}
		\toprule 
		\multirow{2}{*}{$w$}& \multirow{2}{*}{\textbf{Algorithm}} & \multicolumn{4}{c}{$n$} \\
		\cmidrule(l){3-6}
		& & 1K & 5K & 10K & 15K  \\
		\midrule 
		\multirow{2}{*}{3} & \text{KLM} & 22.59  & 270.77 & 1151.86 & 2375.56\\
		& \text{GNN}& 0.017 & 0.040 & 0.073 & 0.104\\
		\midrule 
		\multirow{2}{*}{34} & \text{KLM} & 7.62  & 43.57 & 164.46 & 305.61\\
		& \text{GNN}& 0.020 & 0.074 & 0.145 & 0.223\\
		\bottomrule
	\end{tabular}
\end{table} 

Furthermore, our GNN runs on graphics processing units (GPUs) and thus benefits from accelerated computation. We show running times for the GNN vs.\ KLM ($\epsilon = 0.1$, $\delta = 0.05$) for formulas with ($w=3, 34$, $m$ = 0.75$n$) at every $n$ in Table \ref{tab:runtimes}, and running times for all widths are provided in the appendix. However, we note that KLM and the GNN ran over different hardware (Haswell E5-2640v3 CPU vs.\ P100 GPU, resp.), since they are best suited to their respective devices (CPUs more efficiently handle multiple operations, like sampling, slicing, and comparison, whereas GPUs are efficient for repetitive floating point operations). Hence, these running times are only provided to highlight the scalability of the GNN with increasing formula size, which supports the formal running time expectations of the respective algorithms. Indeed, KLM requires $7.62$ seconds for $w=34, n=1K$, and this rises rapidly to $305.61$s for $w=34, n=15K$, whereas the GNN only needs $0.02$ and $0.223$ seconds, respectively. This is because the GNN takes advantage of limited width to deliver linear scalability, while KLM scales quadratically with $n$ and $m$.

Finally, the GNN does not perform slower at smaller widths as with KLM, as it does not rely on sampling. With KLM, random assignments are replaced when they satisfy a clause, which means that with smaller clause widths, more replacements are made, as clause satisfaction is more likely, and this causes a heavy computational overhead. For example, KLM needs 2375 seconds (about 40 minutes) to run on a formula with $n=15K$ and $w=3$, using $\epsilon = 0.1$ and $\delta = 0.05$, whereas it only requires 306s when $w=34$. By contrast, the GNN requires only 0.104 and 0.223 seconds, respectively. 

\subsection{Discussions: Analyzing the Model}
\begin{figure}[t]
	\centering
	\begin{tikzpicture}[scale=0.9]
	\begin{axis}[
	width=0.5\textwidth,
	tick align=outside,
	tick pos=left,
	ylabel near ticks,
	xlabel near ticks,
	x grid style={white!69.01960784313725!black},
	xlabel={Message Passing Iteration},
	xmin=-0.5, xmax=31.5,
	xtick={1.5,5.5,9.5,13.5,17.5,21.5,25.5,29.5},
	xticklabels={1,2,3,4,5,6,7,8},
	y grid style={white!69.01960784313725!black},
	ylabel={Formulas},
	ymin=-0.5, ymax=20.5,
	ytick={0,1,2,3,4,5,6,7,8,9,10,11,12,13,14,15,16,17,18,19,20},
	yticklabels={1,,3,,5,,7,,9,,11,,13,,15,,17,,19,,21}
	]
	\addplot graphics [includegraphics cmd=\pgfimage,xmin=-0.5, xmax=31.5, ymin=20.5, ymax=-0.5] {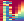};
	\end{axis}
	\end{tikzpicture}
	\hfill
	\caption{GNN estimates over message passing iterations. Red denotes small probability and blue  high probability.}
	\label{fig:probViz}	
\end{figure}
To examine how our model makes predictions, we selected 21 formulas $f_i: i \in [1,21]$ from the \emph{structure} test set with weighted KLM model counts of roughly $\frac{21-i}{20}$. We then ran our GNN model on these formulas and computed the predicted probability at the end of every message passing iteration. Results are visualized in Figure~\ref{fig:probViz}. 
Initially, the network starts with a low estimate. Then, in the first 3 iterations, it accumulates probabilities and hits a ``spike'', which can be mapped to messages from literal nodes reaching the disjunction node. Following this, the network lowers its estimates, before adjusting and refining them in the final iterations. 

Unlike \cite{Selsam-ICLR2019}, where the estimate of satisfiability increases mostly monotonically, our network estimates fluctuate aggressively. A large initial estimate is made, and then reduced and refined. In doing so, the network seems to be initially estimating the naive sum of conjunction probabilities, and subsequently revisiting its estimates as it better captures intersections between conjunctions. This falls in line with our observations, as any understanding of intersections can only occur starting from the third iteration, when the disjunction and conjunction nodes will have passed each other more global information. This also explains the limited performance observed in our ablation study: With just 2 iterations, the system cannot capture conjunction intersections, so can only make naive estimates.

\section{Related Work} 
Weighted \sDNF belongs to the wider family of \WMC problems, which have been extensively studied due to their connection with probabilistic inference. Weighted \sDNF is \sharpP-hard \cite{Valiant79}, so is highly intractable. In fact, Toda proved that the class $\PTime^\sharpP$ contains the entire polynomial hierarchy \cite{Toda-PP89}. Surprisingly, even weighted \sDNF counting on positive, partitioned \DNF formulas with clause width at most~$2$~\cite{Provan-SIAM83} remains \sharpP-hard.

As a result, many methods have been developed to exactly solve or approximate \WMC solutions. One such method is \emph{knowledge compilation (KC)}, where \WMC problems are compiled into a new representation in which they are solved efficiently and exactly. KC pushes computational overhead to a preprocessing phase, but compensates for this by subsequently enabling efficient, linear-time probabilistic inference~\cite{DaMA11}. However, compiled representations can be of exponential size in the worst-case. Hence, KC has limited scalability and robustness to model change, which has motivated research in approximate KC \cite{Lowd10,Friedman18}. Our model training emulates KC preprocessing, but ultimately our model provides approximations and is more scalable. Moreover, it is robust to input changes, as it conceptually handles any formula over a fixed domain of variables and any probability distribution. 

Another important paradigm is to produce approximate solutions to circumvent the intractability of \WMC \cite{Stockmeyer83}.  For the unweighted case (\MC), \emph{hashing-based methods}~\cite{Ermon13,ChakMV13} (see also \cite{CMV16}) produce an approximation with probabilistic guarantees. Importantly, \cite{ChakMV13} also yields an FPRAS when restricted to unweighted DNF; see, e.g.,  \cite{Meel17-DNF}. For hashing methods, approximation-preserving reductions from WMC to MC are known for CNF, but this remains open for the case of DNF \cite{CFMV15}. Hence, \emph{none} of these hashing methods apply to weighted \sDNF.
%\iffalse % This is a moot point considering they don't apply for weighted \sDNF in the first place.
%Furthermore, these methods are at best of the same computational complexity as KLM, and so do not provide any scalability gain for \sDNF. 
%\fi 
Beyond hashing techniques, \emph{loopy belief propagation (LBP)} \cite{Pearl-AAAI82,Murphy-UAI99} has been applied to approximate \WMC. LBP does not provide any guarantees \cite{Weiss-NC00}. Conceptually, our work also uses message passing, but instead learns messages and states so as to best capture the necessary information to relay. It also restricts all outgoing messages from a node to be identical.

Our work builds on recent applications of GNNs \cite{Scarselli09} to a variety of reasoning tasks, such as solving \SAT~\cite{Selsam-ICLR2019} and the traveling salesman problem (TSP)~\cite{Prates-AAAI2019}. There has also been work towards outright learning of inference in probabilistic graphical models \cite{Yoon-ICLRW18}. These works achieve encouraging results, but only on very small instances (i.e., $\sim$40 variables) of their respective problems. Indeed, they struggle to generalize to larger (but still small) instances. This is expected, since SAT and TSP are \NP-complete and are hard to approximate with strong guarantees. Similarly, probabilistic inference in graphical models is \sharpP-hard and remains \NP-hard to approximate (as is weighted \sCNF). Thus, significant work must be done in this direction to reach results of practical use.  In contrast, our work tackles a problem with a known polynomial-time approximation, and learns from a dense dataset of approximate solutions with a very high accuracy at a large scale, and can generalize even further with tolerable loss in performance. To our knowledge, our model is the first proposal that combines reasoning and deep learning, while also scaling to realistic problem instance sizes. 

\section{Summary and Outlook}
We presented Neural\sDNF, a neural-symbolic approach that leverages the traditional KLM approximate weighted \sDNF counter and GNNs to produce weighted \sDNF approximations. This work shows that neural networks can be effectively and efficiently applied to large-scale weighted \sDNF, given sufficiently dense and reliable training data. Therefore, it is particularly useful for query evaluation on large online probabilistic databases, where queries have computational limitations~\cite{CDV-KR16}.

Looking forward, we will analyze the viability of GNNs for other reasoning problems, particularly in light of their expressive power, which could be limiting for problems with less structured graph representations.
%Looking forward, we aim to apply our framework to other problems presenting FPRAS algorithms and a graph formulation. We will also analyze the viability of GNNs in light of their expressive power, which could be limiting for less structured graphs. 
We hope that this work inspires further research leading to less data-dependent neural-symbolic methods, and a greater understanding of neural method performance over challenging problems.

\section{Acknowledgements}

This work was supported by the Alan Turing Institute under the UK EPSRC grant EP/N510129/1, the AXA Research Fund, and by the EPSRC grants EP/R013667/1, EP/L012138/1, and EP/M025268/1. Ralph Abboud is funded by the Oxford-DeepMind Graduate Scholarship and the Alun Hughes Graduate Scholarship. Experiments for this work were conducted on servers provided by the Advanced Research Computing (ARC) cluster administered by the University of Oxford.

\bibliography{library.bib}

\begin{thebibliography}{}

\bibitem[\protect\citeauthoryear{Ba, Kiros, and Hinton}{2016}]{Ba16}
Ba, J.~L.; Kiros, J.~R.; and Hinton, G.~E.
\newblock 2016.
\newblock Layer normalization.
\newblock {\em arXiv preprint arXiv:1607.06450}.

\bibitem[\protect\citeauthoryear{Borgwardt, Ceylan, and
  Lukasiewicz}{2017}]{BCL-AAAI17}
Borgwardt, S.; Ceylan, {\.I}.~{\.I}.; and Lukasiewicz, T.
\newblock 2017.
\newblock Ontology-mediated queries for probabilistic databases.
\newblock In {\em Proc.\ of AAAI}.

\bibitem[\protect\citeauthoryear{Cadoli and Donini}{1997}]{Cadoli97}
Cadoli, M., and Donini, F.
\newblock 1997.
\newblock {A survey on knowledge compilation}.
\newblock {\em AI Communications} 10(3-4).

\bibitem[\protect\citeauthoryear{Ceylan, Darwiche, and {Van den
  Broeck}}{2016}]{CDV-KR16}
Ceylan, {\.I}.~{\.I}.; Darwiche, A.; and {Van den Broeck}, G.
\newblock 2016.
\newblock Open-world probabilistic databases.
\newblock In {\em Proc.\ of KR}.

\bibitem[\protect\citeauthoryear{Chakraborty \bgroup et al\mbox.\egroup
  }{2015}]{CFMV15}
Chakraborty, S.; Fried, D.; Meel, K.~S.; and Vardi, M.~Y.
\newblock 2015.
\newblock {From weighted to unweighted model counting}.
\newblock In {\em Proc.\ of IJCAI}.

\bibitem[\protect\citeauthoryear{Chakraborty, Meel, and Vardi}{2013}]{ChakMV13}
Chakraborty, S.; Meel, K.~S.; and Vardi, M.~Y.
\newblock 2013.
\newblock A scalable approximate model counter.
\newblock In {\em Proc.\ of {CP}}.

\bibitem[\protect\citeauthoryear{Chakraborty, Meel, and Vardi}{2016}]{CMV16}
Chakraborty, S.; Meel, K.~S.; and Vardi, M.~Y.
\newblock 2016.
\newblock Algorithmic improvements in approximate counting for probabilistic
  inference: From linear to logarithmic {SAT} calls.
\newblock In {\em Proc.\ of IJCAI}.

\bibitem[\protect\citeauthoryear{Chung \bgroup et al\mbox.\egroup
  }{2014}]{Chung14}
Chung, J.; Gulcehre, C.; Cho, K.; and Bengio, Y.
\newblock 2014.
\newblock Empirical evaluation of gated recurrent neural networks on sequence
  modeling.
\newblock {\em arXiv preprint arXiv:1412.3555}.

\bibitem[\protect\citeauthoryear{Clevert, Unterthiner, and
  Hochreiter}{2016}]{Djork16}
Clevert, D.; Unterthiner, T.; and Hochreiter, S.
\newblock 2016.
\newblock Fast and accurate deep network learning by exponential linear units
  {(ELUs)}.
\newblock In {\em Proc.\ of ICLR}.

\bibitem[\protect\citeauthoryear{Darwiche and Marquis}{2002}]{DaMA11}
Darwiche, A., and Marquis, P.
\newblock 2002.
\newblock A knowledge compilation map.
\newblock {\em JAIR} 17(1).

\bibitem[\protect\citeauthoryear{{De Raedt}, Kimmig, and
  Toivonen}{2007}]{ProbLog}
{De Raedt}, L.; Kimmig, A.; and Toivonen, H.
\newblock 2007.
\newblock {ProbLog: A probabilistic prolog and its application in link
  discovery}.
\newblock In {\em Proc.\ of IJCAI}.

\bibitem[\protect\citeauthoryear{Domshlak and Hoffmann}{2007}]{Domshlak07}
Domshlak, C., and Hoffmann, J.
\newblock 2007.
\newblock Probabilistic planning via heuristic forward search and weighted
  model counting.
\newblock {\em JAIR} 30(1).

\bibitem[\protect\citeauthoryear{Ermon \bgroup et al\mbox.\egroup
  }{2013}]{Ermon13}
Ermon, S.; Gomes, C.~P.; Sabharwal, A.; and Selman, B.
\newblock 2013.
\newblock Taming the curse of dimensionality: Discrete integration by hashing
  and optimization.
\newblock In {\em Proc.\ of ICML}.

\bibitem[\protect\citeauthoryear{Friedman and Van~den
  Broeck}{2018}]{Friedman18}
Friedman, T., and Van~den Broeck, G.
\newblock 2018.
\newblock Approximate knowledge compilation by online collapsed importance
  sampling.
\newblock In {\em Proc.\ of NeurIPS}.

\bibitem[\protect\citeauthoryear{Gomes, Sabharwal, and Selman}{2009}]{Gomes09}
Gomes, C.~P.; Sabharwal, A.; and Selman, B.
\newblock 2009.
\newblock Model counting.
\newblock In {\em Handbook of Satisfiability}. {IOS} Press.

\bibitem[\protect\citeauthoryear{Gori, Monfardini, and
  Scarselli}{2005}]{Gori2005}
Gori, M.; Monfardini, G.; and Scarselli, F.
\newblock 2005.
\newblock A new model for learning in graph domains.
\newblock In {\em Proc.\ of IJCNN}.

\bibitem[\protect\citeauthoryear{Karp, Luby, and Madras}{1989}]{Karp89}
Karp, R.~M.; Luby, M.; and Madras, N.
\newblock 1989.
\newblock {Monte-Carlo} approximation algorithms for enumeration problems.
\newblock {\em J. Algorithms} 10(3).

\bibitem[\protect\citeauthoryear{Kashima, Tsuda, and
  Inokuchi}{2003}]{Kashima03}
Kashima, H.; Tsuda, K.; and Inokuchi, A.
\newblock 2003.
\newblock Marginalized kernels between labeled graphs.
\newblock In {\em Proc.\ of ICML}.

\bibitem[\protect\citeauthoryear{Kingma and Ba}{2015}]{Kingma-ICLR2014}
Kingma, D.~P., and Ba, J.
\newblock 2015.
\newblock Adam: A method for stochastic optimization.
\newblock In {\em Proc.\ of ICLR}.

\bibitem[\protect\citeauthoryear{Koller and Friedman}{2009}]{Koller-PGM}
Koller, D., and Friedman, N.
\newblock 2009.
\newblock {\em Probabilistic Graphical Models: Principles and Techniques}.
\newblock MIT Press.

\bibitem[\protect\citeauthoryear{Li \bgroup et al\mbox.\egroup }{2016}]{Li15}
Li, Y.; Tarlow, D.; Brockschmidt, M.; and Zemel, R.
\newblock 2016.
\newblock Gated graph sequence neural networks.
\newblock In {\em Proc.\ of ICLR}.

\bibitem[\protect\citeauthoryear{Lowd and Domingos}{2010}]{Lowd10}
Lowd, D., and Domingos, P.~M.
\newblock 2010.
\newblock Approximate inference by compilation to arithmetic circuits.
\newblock In {\em Proc.\ of NIPS}.

\bibitem[\protect\citeauthoryear{Meel, Shrotri, and Vardi}{2017}]{Meel17-DNF}
Meel, K.~S.; Shrotri, A.~A.; and Vardi, M.~Y.
\newblock 2017.
\newblock On hashing-based approaches to approximate {DNF}-counting.
\newblock In {\em Proc. of FSTTCS}.

\bibitem[\protect\citeauthoryear{Meel, Shrotri, and Vardi}{2018}]{Meel18}
Meel, K.; Shrotri, A.; and Vardi, M.
\newblock 2018.
\newblock Not all {FPRAS}s are equal: {D}emystifying {FPRAS}s for
  {DNF}-counting.
\newblock {\em Constraints}.

\bibitem[\protect\citeauthoryear{Morris \bgroup et al\mbox.\egroup
  }{2019}]{GroheAAAI19}
Morris, C.; Ritzert, M.; Fey, M.; Hamilton, W.~L.; Lenssen, J.~E.; Rattan, G.;
  and Grohe, M.
\newblock 2019.
\newblock Weisfeiler and {L}eman go neural: Higher-order graph neural networks.
\newblock In {\em Proc.\ of AAAI}.

\bibitem[\protect\citeauthoryear{Murphy, Weiss, and
  Jordan}{1999}]{Murphy-UAI99}
Murphy, K.~P.; Weiss, Y.; and Jordan, M.~I.
\newblock 1999.
\newblock Loopy belief propagation for approximate inference: An empirical
  study.
\newblock In {\em Proc.\ of {UAI}}.

\bibitem[\protect\citeauthoryear{Pearl}{1982}]{Pearl-AAAI82}
Pearl, J.
\newblock 1982.
\newblock Reverend {B}ayes on inference engines: {A} distributed hierarchical
  approach.
\newblock In {\em Proc.\ of {AAAI}}.

\bibitem[\protect\citeauthoryear{Prates \bgroup et al\mbox.\egroup
  }{2019}]{Prates-AAAI2019}
Prates, M. O.~R.; Avelar, P. H.~C.; Lemos, H.; Lamb, L.; and Vardi, M.
\newblock 2019.
\newblock Learning to solve {NP}-complete problems - {A} graph neural network
  for the decision {TSP}.
\newblock In {\em Proc.\ of AAAI}.

\bibitem[\protect\citeauthoryear{Provan and Ball}{1983}]{Provan-SIAM83}
Provan, J.~S., and Ball, M.~O.
\newblock 1983.
\newblock The complexity of counting cuts and of computing the probability that
  a graph is connected.
\newblock {\em SIAM} 12(4).

\bibitem[\protect\citeauthoryear{Scarselli \bgroup et al\mbox.\egroup
  }{2009}]{Scarselli09}
Scarselli, F.; Gori, M.; Tsoi, A.~C.; Hagenbuchner, M.; and Monfardini, G.
\newblock 2009.
\newblock The graph neural network model.
\newblock {\em IEEE Transactions on Neural Networks} 20(1).

\bibitem[\protect\citeauthoryear{Selman and Kautz}{1996}]{SeKa96}
Selman, B., and Kautz, H.
\newblock 1996.
\newblock {Knowledge compilation and theory approximation}.
\newblock {\em JACM} 43(2).

\bibitem[\protect\citeauthoryear{Selsam \bgroup et al\mbox.\egroup
  }{2019}]{Selsam-ICLR2019}
Selsam, D.; Lamm, M.; B{\"u}nz, B.; Liang, P.; de~Moura, L.; and Dill, D.~L.
\newblock 2019.
\newblock Learning a {SAT} solver from single-bit supervision.
\newblock In {\em Proc.\ of ICLR}.

\bibitem[\protect\citeauthoryear{Stockmeyer}{1983}]{Stockmeyer83}
Stockmeyer, L.
\newblock 1983.
\newblock The complexity of approximate counting.
\newblock In {\em Proc.\ of STOC}.
\newblock ACM.

\bibitem[\protect\citeauthoryear{Suciu \bgroup et al\mbox.\egroup
  }{2011}]{Suciu-PDBs}
Suciu, D.; Olteanu, D.; R{\'{e}}, C.; and Koch, C.
\newblock 2011.
\newblock {\em {Probabilistic Databases}}, volume~3.
\newblock Morgan \& Claypool.

\bibitem[\protect\citeauthoryear{Toda}{1989}]{Toda-PP89}
Toda, S.
\newblock 1989.
\newblock On the computational power of {PP and +P}.
\newblock In {\em Proc.\ of FOCS}.

\bibitem[\protect\citeauthoryear{Valiant}{1979}]{Valiant79}
Valiant, L.~G.
\newblock 1979.
\newblock {The complexity of computing the permanent}.
\newblock {\em TCS} 8(2).

\bibitem[\protect\citeauthoryear{Wang \bgroup et al\mbox.\egroup
  }{2014}]{Wang14}
Wang, Z.; Zhang, J.; Feng, J.; and Chen, Z.
\newblock 2014.
\newblock Knowledge graph embedding by translating on hyperplanes.
\newblock In {\em Proc.\ of AAAI}.

\bibitem[\protect\citeauthoryear{Weiss}{2000}]{Weiss-NC00}
Weiss, Y.
\newblock 2000.
\newblock Correctness of local probability propagation in graphical models with
  loops.
\newblock {\em Neural Computation} 12(1).

\bibitem[\protect\citeauthoryear{Xu \bgroup et al\mbox.\egroup
  }{2019}]{Keyulu18}
Xu, K.; Hu, W.; Leskovec, J.; and Jegelka, S.
\newblock 2019.
\newblock How powerful are graph neural networks?
\newblock In {\em Proc.\ of ICLR}.

\bibitem[\protect\citeauthoryear{Yoon \bgroup et al\mbox.\egroup
  }{2018}]{Yoon-ICLRW18}
Yoon, K.; Liao, R.; Xiong, Y.; Zhang, L.; Fetaya, E.; Urtasun, R.; Zemel,
  R.~S.; and Pitkow, X.
\newblock 2018.
\newblock Inference in probabilistic graphical models by graph neural networks.
\newblock In {\em Workshop Proc. of {ICLR}}.

\end{thebibliography}
\bibliographystyle{aaai}

\cleardoublepage
\appendix
\section{Details of Data Generation}
\label{app:datagen}
 \subsection{Standard Generation Procedure}
To generate data, we develop a comprehensive randomized generation procedure, which takes as input the target number of variables $n$, the target number of clauses $m$, and minimum and maximum bounds $\minw$ and $\maxw$ on clause widths $w$ within the formula. In the paper, we use fixed-width clauses, hence, set $\minw = \maxw = w$. Initially, the procedure randomly generates $m$ clause widths using a uniform distribution bounded between $\minw$ and $\maxw$ inclusive. It then computes their sum, which we call the \emph{slot count} $s$, and continues generation only if ${s \geq n}$. It then allocates the $s$ slots to the $n$ variables, such that every variable is allocated at least one slot, ensuring all variables appear in the generated formula. This is equivalent to the combinatorial problem of putting $k$ balls into $n$ boxes such that no boxes are empty. 

Once all variable allocations are determined, all variables are sorted in decreasing allocation order and then assigned to clauses in that order. This ensures that more prominent variables, which appear more in the formula, are accommodated first, when more empty slots are available, thus maximizing the likelihood of generation success. In this assignment phase, a variable having $s$ slot allocations will be assigned to $s$ clauses by randomly, selecting these clauses from the $m$ total clauses without replacement. This ensures no variable is assigned to the same clause twice, to prevent redundancy. Further heuristics are also added to this mechanism to prioritize selecting clauses with more empty slots at this phase, so that all clauses are filled in a uniform fashion. At the end of variable assignment to clauses, all variable instances are individually randomized to be positive or negative literals. % Last sentence added following rebuttal

Probabilities for variables are chosen uniformly at random. In our experiments, we use 4 distributions for every formula in our training and evaluation sets, such that one distribution is randomly generated, and the other 3 distributions are quarter increments of the random distribution modulo 1. For example, if a variable $v_1$ is assigned probability $0.1$ for the random distribution, it will have probabilities $0.35$, $0.6$, and $0.85$, respectively, in the 3 other distributions. This ensures that we produce formulas with model counts covering the entire $[0,1]$ range as evenly as possible, so as to have more representative training and testing data.

\subsection{Adding Non-uniformity to Generation}
In the standard generation procedure presented earlier, every variable appears between 1 and $m$ times in a formula, and $\frac{S}{n}$ times in expectation. However, it is highly unlikely, by the Chernoff bound, to produce formulas with high dependence on a small subset of variables, i.e., with some variables appearing far more frequently than others and across a majority of clauses. Thus, with very high probability, formulas generated according to this procedure will have their weighted model count depend almost exclusively on clause widths and $m$, with individual variables having very little impact on the model count. This is highly undesirable, as it prevents the network from learning the contributions of individual variables, and encourages overfitting to higher-level structural details of the formula, namely, $w$ and $m$.

To tackle this, we introduce two new variables ${q , r \in [0,1]}$. These variables create a set of  $p = q\cdot n$ \emph{privileged} variables, which will appear far more frequently than their non-privileged counterparts and impact the model counts more severely. These variables are exclusively assigned $r\cdot e$ slots at random, where $e$ is the  \emph{excess} slots $e = s - n$. The remaining $s - e\cdot r$ slots are then subsequently allocated to all variables (including privileged ones) using the standard generation procedure, and assignment to clauses is done analogously afterwards. Therefore, the expected number of allocations for a privileged variable $v_{priv}$ given non-zero values of $q$ and $r$ is: \[1 + \frac{r\cdot e}{q\cdot n} + \frac{(1 - r)\cdot e}{n} = 1 + \frac{e(q(1-r) + r)}{q\cdot n}.\]To further enforce dependence on privileged variables, all corresponding privileged literals for a privileged variable are randomised together, such that they all are unanimously positive or negative. This makes that all clauses require the same assignment of the privileged variable, and that no mutually exclusive clauses are created, which would greatly increase formula counts and reduce variability. Therefore,  literals of privileged variables are all set to the same sign to give them more impact on the model count 

\subsection{Generating Formulas for Experiments}
When generating formulas for the experimental data sets, we set $q = 0$ and $r = 0$ with probability 0.5, and therefore generate 50\% of our formulas without privileged variables. For the remaining 50\%, we sample $q$ from an exponential distribution with $\lambda = 1$, take the remainder of this value modulo $\frac{\log{(n)}}{n}$, and round this value up to the nearest multiple of $\frac{1}{n}$. This privileged variable selection process is highly selective. In fact, this process becomes increasingly selective as $n$ increases, since $\lim_{n\to\infty} \frac{\log{(n)}}{n} = 0$. As a result, privileged variables have a strong effect on the formula model count. With q set, we finally, set $r$ as being the value for which generation can succeed (i.e., no privileged variable gets allocated more than the number of clauses) with probability at least 0.5 by a one-sided Chebyshev bound.

\section{Further Experiments}
\label{app:exp}
\subsection{Ablation Testing}

To evaluate the role of message passing for model performance, we ran the same experimental protocol as in Section \ref{sec:exp} using only $T=2$ iterations for both training and testing. With 2 iterations, only one full pass through the 3 network layers is possible: in the first iteration, literal node messages update all conjunction node states, and in the second iteration, these nodes provide a meaningful update to the disjunction node. With this configuration, we observed a large drop in system performance, both in terms of structure and size generalization. The results of this experiment are provided in Tables \ref{tab:ablationTrTest} and \ref{tab:ablationGen}. From this, we learn that message passing is essential to model performance. Indeed, the model cannot deliver a reliable estimate within just 2 iterations, as it cannot learn about intersections between clauses, thereby limiting it to naive guesses based on disjoint unions. This is also evidenced in Figure \ref{fig:probViz}, where this phenomenon occurs within the first 2 iterations of the 8-iteration set.
\begin{table}[t]
		\centering
		\caption{Accuracy (\%) on all thresholds on training and structure evaluation datasets for ablation study ({$T=2$}).} 
		\label{tab:ablationTrTest} 
\begin{tabular}{lcccc}
			\toprule 
			\multirow{2}{*}{\textbf{Evaluation Data}} & \multicolumn{4}{c}{\textbf{Thresholds}} \\
		\cmidrule(l){2-5}
		& 0.02 & 0.05 & 0.10 & 0.15  \\
			\midrule 
			\text{Training Set} & 71.84 & 82.91 & 91.62 & 95.18 \\
			\text{Test Set} & 72.08 & 82.80 & 91.36 & 94.97 \\
			\bottomrule
		\end{tabular}
\end{table}
\begin{table}
\centering
		\caption{Accuracy (\%) on all thresholds on size generalization datasets for ablation study ($T=2$).} 
		\label{tab:ablationGen} 
\begin{tabular}{lcccc}
			\toprule 
			\multirow{2}{*}{$n$} & \multicolumn{4}{c}{\textbf{Thresholds}} \\
		\cmidrule(l){2-5}
		& 0.02 & 0.05 & 0.10 & 0.15  \\
			\midrule 
			\textbf{10K} & 71.84 & 76.72 & 84.20 & 89.94 \\
			\textbf{15K} & 66.37 & 69.83 & 79.31 & 87.93 \\
			\bottomrule
		\end{tabular}
\end{table}

\subsection{Experiments with More Iterations}
In addition to our ablation study, we also train and run our system using ${T=32}$ message passing iterations. With this many iterations, the system is expected to build a more comprehensive understanding of connections between different components of the graph, and so should perform better. However, we found that this was not the case. In terms of structure generalization (Table \ref{tab:testSet32}), the system achieves similar performance compared with $T=8$, but performs significantly worse in terms of size generalization (Table \ref{tab:sizeGen32}). This shows that the system \emph{overfits} with too many iterations, such that it learns a message passing optimized for its training set, but cannot generalize to larger formulas. 
\begin{table}[t]
	\centering
	\caption{Accuracy (\%) on all thresholds on training and structure evaluation datasets for ${T=32}$.} 
	\label{tab:testSet32} 
	\begin{tabular}{lccccc}
			\toprule 
			\multirow{2}{*}{\textbf{Evaluation Data}} & \multicolumn{4}{c}{\textbf{Thresholds}} \\
			\cmidrule(r){2-5}
			& 0.02 & 0.05 & 0.10 & 0.15  \\
			\midrule 
			\text{Training Set} & 87.68 & 98.22 & 99.87 & 99.99 \\
			\text{Test Set} & 87.32 & 98.02 & 99.83 & 99.95 \\
			\bottomrule
		\end{tabular}
\end{table}

\begin{table}[t]
	\centering
		\caption{Accuracy (\%) on all thresholds on size generalization datasets for ${T=32}$.} 
	\begin{tabular}{lcccc}
		\toprule % <-- Toprule here
		\multirow{2}{*}{\textbf{$n$}} & \multicolumn{4}{c}{\textbf{Thresholds}} \\
		\cmidrule(l){2-5}
		& 0.02 & 0.05 & 0.10 & 0.15  \\
		\midrule % <-- Midrule here
		\textbf{10K} & 74.43 & 83.91 & 93.67 & 96.26 \\
		\textbf{15K} & 68.97 & 80.17 & 87.93 & 92.24\\
		\bottomrule
	\end{tabular}
	\label{tab:sizeGen32} 
\end{table}
\subsection{Results on Other Datasets}
We generated two new synthetic datasets of identical size and composition to the original structure evaluation set. The first dataset is entirely based on \emph{our} generator, such that privileged variables are always used during generation (in the main experiments, this is only used 50\% of the time). As a result, this dataset consists entirely of formulas where a small subset of variables have a major effect on the overall weighted model count. We refer to this dataset as \emph{Fully Privileged}. 

The second dataset, on the other hand, is generated using a random generator similar to the one used in \cite{Meel18}, where variables and clauses are generated and allocated uniformly at random, and we call it  the \emph{Meel et al.} dataset.
We evaluated our model on both these sets, and show results in Table \ref{tab:newsets}. Our system performs very well on the ``privileged'' data set, in close proximity to original test set performance, and actually performs better on the fully random dataset. This is not surprising, as \cite{Meel18} allocates variables uniformly, thus producing formulas with model counts that can be statistically approximated more easily. These results highlight the robustness of our GNN, and also show the quality of our generation procedure. 
\begin{table}[h]
\centering
\caption{Results over other synthetic datasets.}
		\label{tab:newsets}
		\begin{tabular}{lcccc}
			\toprule 
			\multirow{2}{*}{\textbf{Dataset}} & \multicolumn{4}{c}{\textbf{Thresholds}} \\
		\cmidrule(l){2-5}
		& 0.02 & 0.05 & 0.10 & 0.15  \\
			\midrule 
			\text{Fully Privileged} & 86.33 & 98.88 & 99.96 & 99.99 \\			
			\text{Meel et al.} & 88.26 & 98.67 & 99.98 & 100.0 \\		
			\bottomrule
		\end{tabular}
\end{table} 

\section{Complexity of Algorithms}
\label{app:complx}
Let $n$ be the number of variables in a DNF formula, $m$ be the number of clauses, and $\bar{w}$ be the average clause width. We show that our tool has average-case complexity $O(m\bar{w})$ and is therefore more efficient than KLM, which has complexity $O\big(nm \epsilon^{-2}\log(\frac{1}{\delta})\big)$, since, in practice, $\bar{w} << n$:
\begin{itemize}
\item In a message passing iteration, $2\sum_{i=1}^m w_i$ messages are passed between \emph{literal} layer nodes and \emph{conjunction} layer nodes, where $w_i$ denotes the width of clause $i$. $2\sum_{i=1}^m w_i$ can be rewritten as $2m\bar{w}$, where $\bar{w}=\frac{1}{m}\sum_{i=1}^m w_i$. 

Furthermore,  $2n$ messages pass between \emph{literal} nodes and their negations, and $2m$ messages pass between \emph{conjunction} nodes and the \emph{disjunction} node. All messages are summed at their respective destinations, thus yielding as many additions as messages passed. Following this, $2n+m+1$ constant-time state updates are made. Since every variable is assumed to appear in a clause, we always have that $n \leq mw$. As a result, the average-case complexity of a message passing iteration is $O(m\bar{w})$. Finally, since the number of message passing iterations $T$ is fixed, we deduce that the average-case GNN complexity is also $O(m\bar{w})$, which is significantly better than KLM's $O\big(nm \epsilon^{-2}\log(\frac{1}{\delta})\big)$, particularly since ${\bar{w} << n}$, in practice. 

\item Even in the worst-case where all clauses have width $n$, ${2nm}$ messages are passed between \emph{literals} nodes and \emph{conjunctions} nodes, and the GNN therefore has complexity ${O(nm)}$, which is asymptotically identical to KLM. 

\item In the best-case, when clause width is upper-bounded by a constant, the number of messages passed becomes ${O(n+m)}$. And since state updates are also $O{(n+m)}$, the complexity of an iteration therefore reduces to ${O(n+m)}$, allowing for \emph{linear-time} estimation of weighted DNF counts. 
\end{itemize}

\section{Algorithm Running Times}
\label{app:runtime}
KLM ran on a single Haswell E5-2640v3 CPU, which has a clock speed of 2.60GHz. This CPU was part of a node with has 64 GB of memory.  Our GNN ran on a Tesla P100 GPU, which has 12 GB of on-device memory. Running times for both algorithms were measured across 50 runs and averaged to produce the results in Table \ref{tab:allruntimes}.
\begin{table}[t] 
	\centering
	\caption{Runtimes (s) for KLM, GNN by number of variables ($n$) and width ($w$)  for $m$ = 0.75$n$.}
	\label{tab:allruntimes}
	\begin{tabular}{lccccc}
		\toprule 
		\multirow{2}{*}{$w$}& \multirow{2}{*}{\textbf{Algorithm}} & \multicolumn{4}{c}{\textbf{$n$}} \\
		\cmidrule(l){3-6}
		& & 1K & 5K & 10K & 15K  \\
		\midrule 
		\multirow{2}{*}{$3$} & \text{KLM} & 22.59  & 270.77 & 1151.86 & 2375.56\\
		& \text{GNN}& 0.017 & 0.040 & 0.073 & 0.104\\
		\midrule 
		\multirow{2}{*}{$5$} & \text{KLM} & 15.32  & 145.47 & 608.13 & 1298.38\\
		& \text{GNN}& 0.017 & 0.042 & 0.077 & 0.110\\
		\midrule 
		\multirow{2}{*}{$8$} & \text{KLM} & 9.32  & 81.39 & 322.56 & 689.50\\
		& \text{GNN}& 0.017 & 0.045 & 0.083 & 0.121\\
		\midrule 
		\multirow{2}{*}{$13$} & \text{KLM} & 7.07  & 40.58 & 158.09 & 299.26\\
		& \text{GNN}& 0.017 & 0.050 & 0.094 & 0.138\\
		\midrule 
		\multirow{2}{*}{$21$} & \text{KLM} & 7.26  & 41.28 & 157.85 & 293.83\\
		& \text{GNN}& 0.018 & 0.058 & 0.113 & 0.167\\
		\midrule 
		\multirow{2}{*}{$34$} & \text{KLM} & 7.62  & 43.57 & 164.46 & 305.61\\
		& \text{GNN}& 0.020 & 0.074 & 0.145 & 0.223\\
		\bottomrule
	\end{tabular}
\end{table}

\end{document}